\pdfoutput=1

\documentclass[11pt]{article}

\usepackage{latex/acl}

\usepackage{times}
\usepackage{latexsym}

\usepackage[T1]{fontenc}

\usepackage[utf8]{inputenc}

\usepackage{microtype}

\usepackage{inconsolata}

\usepackage{tabularray}
\usepackage{longtable}
\usepackage{xcolor}
\usepackage{colortbl}
\usepackage{multicol}
\usepackage{booktabs}
\usepackage{makecell}
\usepackage{amsmath, amssymb}
\usepackage{multirow}
\usepackage{mathtools}
\usepackage{enumitem}
\usepackage{subcaption}
\usepackage{float}
\usepackage{graphicx}
\usepackage{caption} 
\usepackage{upgreek}
\usepackage{seqsplit}
\usepackage{color,soul}
\usepackage{arydshln}

\usepackage{amsmath,amsfonts,bm}

\def\eqref#1{equation~\ref{#1}}

\def\1{\bm{1}}

\def\vc{{\bm{c}}}

\def\vq{{\bm{q}}}

\def\vx{{\bm{x}}}
\def\vy{{\bm{y}}}

\DeclareMathAlphabet{\mathsfit}{\encodingdefault}{\sfdefault}{m}{sl}
\SetMathAlphabet{\mathsfit}{bold}{\encodingdefault}{\sfdefault}{bx}{n}

\usepackage{caption}
\usepackage{subcaption}
\title{ Retrieval-Augmented Data Augmentation for Low-Resource Domain Tasks }

\author{
    Minju Seo\thanks{* Equal contribution} \:\;
    Jinheon Baek$^*$ \:\;
    James Thorne \:\;
    Sung Ju Hwang \\
    KAIST \\
    \texttt{\{minjuseo, jinheon.baek, thorne, sjhwang82\}@kaist.ac.kr}
}

\begin{document}
\maketitle
\begin{abstract}
Despite large successes of recent language models on diverse tasks, they suffer from severe performance degeneration in low-resource settings with limited training data available. Many existing works tackle this problem by generating synthetic data from the training data and then training models on them, recently using Large Language Models (LLMs). However, in low-resource settings, the amount of seed data samples to use for data augmentation is very small, which makes generated samples suboptimal and less diverse. To tackle this challenge, we propose a novel method that augments training data by incorporating a wealth of examples from other datasets, along with the given training data. Specifically, we first retrieve the relevant instances from other datasets, such as their input-output pairs or contexts, based on their similarities with the given seed data, and then prompt LLMs to generate new samples with the contextual information within and across the original and retrieved samples. This approach can ensure that the generated data is not only relevant but also more diverse than what could be achieved using the limited seed data alone. We validate our proposed Retrieval-Augmented Data Augmentation (RADA) framework on multiple datasets under low-resource settings of training and test-time data augmentation scenarios, on which it outperforms existing LLM-powered data augmentation baselines.
\end{abstract}
\section{Introduction}

\begin{figure*}[t!]
    \vspace{-0.125in}
    \begin{minipage}{0.815\textwidth}
    \centering
    \includegraphics[width=1.0\linewidth]{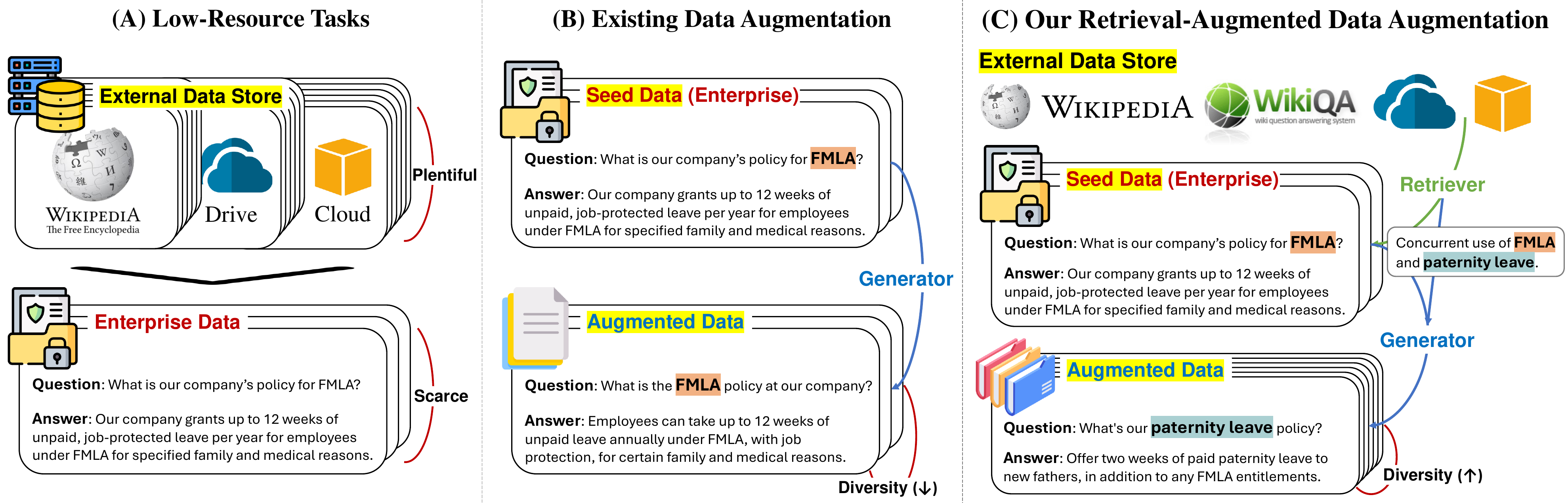}
    \end{minipage}
    \hfill
    \begin{minipage}{0.175\textwidth}
    \includegraphics[width=1\linewidth]{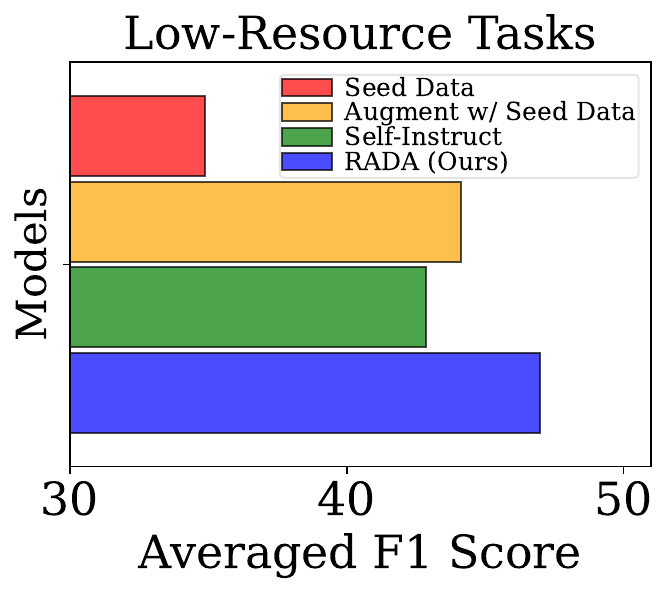}
    \vspace{0.025in}
    \includegraphics[width=1\linewidth]{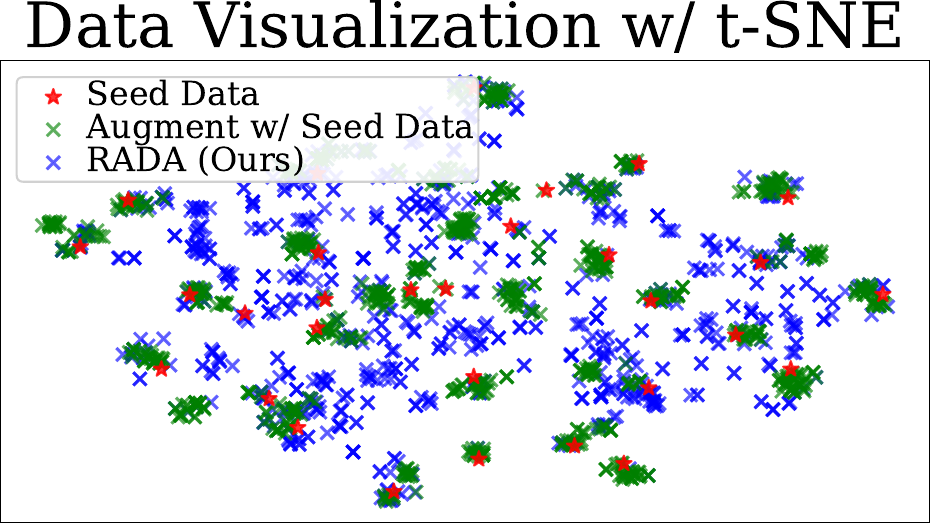}
    \end{minipage}
    \vspace{-0.125in}
    \caption{\textbf{(A) Low-Resource Tasks} refer to problems (usually on the specific domains) where there is a limited amount of data available. \textbf{(B) Existing Data Augmentation} approaches expand the seed data with itself (policy for FMLA), which results in the limited diversity of the generated data samples (the same FMLA policy). \textbf{(C) Our Retrieval-Augmented Data Augmentation (RADA)} framework generates the new data with the external context (concurrent usage of FMLA and paternity leave), retrieved from the external datasets, along with the seed data, yielding more diverse and useful samples (paternity leave). \textbf{(Upper Right:)} Our RADA outperforms existing data augmentation methods, demonstrating the quality of generated samples. \textbf{(Lower Right:)} The generated data samples from RADA are more diverse than existing data augmentation, based on the t-SNE visualization.}
    \label{fig:concept}
    \vspace{-0.175in}
\end{figure*}

Recent advances in language models~\cite{GPT-3, Llama2, GPT-4, Gemini}, which are trained on general text corpora, have achieved numerous successes across various natural language tasks. The common practice to further enhance their performances is to perform fine-tuning on task-specific datasets, which has been proven substantially effective regardless of model sizes~\cite{false/promise, LLM/finetuning}. However, the efficacy of this fine-tuning is closely tied to the volume and quality of the data available for training. Meanwhile, in real-world scenarios, particularly in specific domains, there is often a scarcity of training instances. For example, at the beginning of a pandemic such as COVID-19, there are only a few limited training instances to fine-tune language models, despite an urgent need for tasks, such as question answering~\cite{covidqa} (Figure~\ref{fig:concept}, (A)). Yet, the manual annotation of additional training samples is costly and time-consuming, which may require domain experts. 

To address this challenge, various approaches have been proposed to augment the training data automatically. These methods typically range from altering the texts of existing training samples~\cite{DA/rule/1, DA/rule/2} to leveraging generative models to produce new instances for training based on initial seed samples~\cite{DA/model/1, DA/model/2, DA/model/3}. Also, many recent approaches have leveraged the capability of LLMs for data augmentation based on prompting, which eliminates the burden of performing task-specific training~\cite{DA/LLM/1, DA/LLM/2, DA/LLM/3}. In particular, \citet{evoprompting} has utilized the diverse prompting strategies to create a broader set of instances. However, in low-resource environments where only a limited number of training instances are available, generating new data from these minimal seed samples results in poor diversity and variation (See Figure~\ref{fig:concept}, (B)). We note that a very recent approach attempts to overcome this by iteratively including generated samples as seed data for further data generation~\cite{self_instruct}. However, this approach is still ill-suited, which is not only constrained by the limited diversity of the initial seed data but also vulnerable to recursively diminishing the quality of subsequent augmentations due to the potential low-quality of prior augmentations.

Despite the limited seed data in low-resource settings, there is an abundance of examples and resources accumulated in existing data pools, which can be utilized for data augmentation. Moreover, by leveraging the contextual understanding capabilities of LLMs, we can effectively utilize a mixture of samples drawn from the initial seed data, other datasets, or a combination of both. This can enable the synthesis of new samples, which mirror the characteristics of the original seed data while being diverse, without necessitating additional training.

However, not all samples from external datasets are useful for data augmentation, as most of them may not align with the characteristics of the seed data. Thus, inspired by the motivation to use external data instances while overcoming the problem of many of their irrelevancies, in this work, we propose a novel LLM-powered Retrieval-Augmented Data Augmentation (RADA) framework (See Figure~\ref{fig:concept}, (C)). Specifically, the input of our data augmentation approach consists of in-context examples containing example instances, along with a target context that elicits a new sample generation. To be more specific, for open-domain question answering, which aims to answer a question based on information in a document, a sequence of multiple triplets of the document, question, and answer is used for in-context, while the target context is the document from which new question-answer pairs are generated. Then, our RADA flexibly employs multiple retrieval strategies to construct these in-context and target-context with samples from both original and external datasets, enabling diverse data augmentation, unlike the conventional approaches that rely solely on the initial seed data.

We validate the effectiveness of RADA in augmenting low-resource datasets on multiple domain-specific datasets, where we consider both the training and test-time data augmentation scenarios. The experimental results show that RADA consistently surpasses several LLM-powered data augmentation baselines on all datasets. In addition, a key finding from our analyses is the dual benefit offered by our RADA: the incorporation of external data sources enhances the diversity of the generated instances, while the retrieval mechanism ensures maintaining their semantic alignment with the initial seed data.

Our findings and contributions are threefolds:
\vspace{-0.125in}
\begin{itemize}[itemsep=0.0mm, parsep=1pt]
    \item We point out the limitation of existing data augmentation approaches that rely on initial seed data alone, leading to a lack of diversity.
    \item We introduce a novel retrieval-augmented data augmentation framework, which performs retrieval over external data sources to generate diverse data based on information within and across the original and retrieved samples.
    \item We validate our RADA in augmenting data on low-resource settings with training and test-time scenarios, demonstrating its efficacy in generating the diverse and high-quality data.
\end{itemize}
\section{Related Work}

\subsection{Large Language Models}

Large Language Models (LLMs), which are trained on vast amounts of textual corpora with multiple training strategies along with a large number of parameters, have demonstrated remarkable capability of handling diverse language tasks~\cite{GPT-3, Llama2, GPT-4, Gemini}. A notable feature of these models is their ability to perform in-context learning, which means they can understand and learn from examples or instructions provided in the input and then adapt their responses based on this information, without requiring explicit retraining for each specific task~\cite{GPT-3, CoT, MetaICL, in-context/ssl}. Due to its simplicity yet effectiveness and versatileness across diverse tasks, several approaches have been introduced to improve the quality of the LLM context. To mention a few, \citet{z_icl} constructs pseudo-demonstrations, when examples in the context are not available, by retrieving the relevant instances from the external corpus based on their similarities with the input query. Similarly, \citet{incontext_ralm} and \citet{KAPING} augment LLMs by prepending the relevant documents or facts retrieved from the external corpus in their input context, to subsequently improve the factuality of LLM responses. Lastly, \citet{adapt_incontext} targets adapting LLMs with in-context examples (which are adaptively retrieved) for domain adaptation. However, existing works do not focus on augmenting the data based on the retrieval of its relevant samples from other datasets, based on in-context learning of LLMs.

\subsection{Data Augmentation} 
Despite the notable successes of LLMs, their performance significantly deteriorates in low-resource settings, particularly for domain-specific environments where the data available for training is very scarce (for instance, in the case of emerging events like novel viruses) or, in certain cases, completely unavailable (such as in privacy-sensitive enterprise contexts)~\cite{LLM/domain/1, LLM/domain/2, LLM/domain/3}. Further, they are less likely to be trained with ones similar to these specialized data, leading to constrained capability in handling them. To address this challenge, numerous studies have proposed to expand the original seed data with various data augmentation techniques~\cite{DA/survery/1, DA/survery/2}. Early works utilized token-level perturbation approaches, which either alter texts~\cite{DA/rule/1, DA/rule/2} or interpolate them~\cite{DA/mixup/1, DA/mixup/2}. Recent studies have shifted the focus towards utilizing the capability of generative language models, since they may internalize the useful knowledge to generate samples relevant to the seed data. Previous works on this line trained relatively smaller language models, based on the input-output pairs of the seed data to generate new outputs from the input variants~\cite{DA/model/1, DA/model/2, DA/model/3}. Also, more recent works have used scaled-up versions of language models (called LLMs), which have much greater capability in generating high-quality data (sometimes surpassing human-level performances) without requiring task-specific training~\cite{DA/LLM/1, DA/LLM/2, DA/LLM/3}. Specifically, in information retrieval, some studies have generated synthetic queries with LLMs, to match the unlabeled documents with them~\cite{inpars, promptagator, udapdr}. Similarly, some other studies have proposed LLM-powered data augmentation methods for specific down-stream tasks, such as text classification~\cite{auggpt}, reading comprehension~\cite{can_mrc_low}, or multi-hop question answering~\cite{efficient_multihop}. This trend also goes to empowering the collection of instruction-tuning and alignment datasets for LLM training, which expands actual data samples with synthetic samples generated from LLMs themselves~\cite{unnatural, self_instruct, harnessing_goliath, self_alignment_backtranslation}. However, in the low-resource setting, the seed examples available to use for data augmentation are extremely scarce, which may result in suboptimal quality and limited diversity of the generated data. In this work, we propose to overcome this limitation by augmenting the data generation process with retrieval from larger external samples.

\section{Methodology}

In this section, we present a Retrieval-Augmented Data Augmentation (RADA) framework.

\subsection{Problem Statement}

We begin with introducing the problem of domain-specific tasks under low-resource settings, followed by describing LLMs for data augmentation.

\paragraph{Low-Resource Domain-Specific Tasks}

Before explaining the low-resource tasks that we focus on, we provide a definition of conventional natural language tasks. Formally, they aim to predict a label $\vy$ given an input $\vx$, where $\vx$ and $\vy$ are comprised of a sequence of tokens: $\vx = [x_1, x_2, ..., x_{|\vx|}]$ and $\vy = [y_1, y_2, ..., y_{|\vy|}]$. Then, the training data $\mathcal{D}$ can be represented as an aggregation of input-output pairs: $\mathcal{D} = \left\{ (\vx_i, \vy_i) \right\}_{i=1}^{N}$ where its size $N$ can vary widely from just a few dozens to several millions. 

In this work, we target handling challenging scenarios where $N$ is notably small, usually referred to as low-resource settings. These settings are particularly prevalent in domain-specific tasks (within legal, medical, or technical fields), where the availability of labeled data is inherently limited due to the specialized nature of the domain or the scarcity of domain experts for annotation; however, its quality and size are crucial to train performant models.

\paragraph{LLMs for Data Augmentation}

A typical way to handle the low-resource domain tasks is to expand the training data $\mathcal{D}$ with data augmentation techniques, which has been recently powered by LLMs due to their strong text-generation capabilities. Formally, let us first describe the LLM as a model parameterized by $\theta$, which takes the input $\vx$ and then generates the output $\vy$, represented as follows: $\vy = \texttt{LLM}_{\theta}(\vx)$. Here, $\theta$ is trained with massive text corpora with several training strategies and, after them, it usually remains fixed due to the costs of further training. Also, $\vx$ can be any form of text, referred to as the prompt, which includes task-dependent instructions and contexts (such as demonstrations), to guide LLMs in generating outputs that align with the user's intent, which is data augmentation in our work, discussed below.

The primary goal of data augmentation is to expand the diversity and amount of data $\mathcal{D}$ available for model training (and for testing in certain use cases such as test-time adaption), without manually collecting the new data, for tackling specific tasks especially on low-resource domains. Formally, this data augmentation process can be represented as follows: $\mathcal{D}' = f(\mathcal{D})$, where $f$ is the model (or technique) designed to generate new input-output pairs $(\vx', \vy')$ for the augmented dataset $\mathcal{D}'$, which is achieved by leveraging the underlying patterns, contexts, and knowledge existing in seed data $\mathcal{D}$. However, while there have been great successes in advancing the augmentation methods $f$ in several different ways, for example, training the generative models or further prompting LLMs with the given original data, they mainly focus on expanding the original data $\mathcal{D}$ with itself. On the other hand, we can potentially incorporate any abundant sources of information easily available at hand, which could introduce greater diversity and quality in generating the samples for data augmentation. In addition, especially in low-resource settings, the available data to use as a source for expansion is largely scarce, which poses a significant challenge as the augmentation method $f$ is operationalized with only limited samples, leading to the generation of samples that may lack the desired diversity and quality.

\subsection{Retrieval-Augmented Data Augmentation}

In this work, to tackle the aforementioned drawbacks of the existing data augmentation approaches that are limited by the given dataset, we propose a novel data augmentation approach (from a different angle), that leverages available external datasets.

\paragraph{Data Generation with External Resources} 
We redefine the concept of previous data augmentation to incorporate leveraging samples from external resources, represented as follows: $\mathcal{D}' = f(\mathcal{D}, \mathcal{C})$ where $\mathcal{C}$ is an external data store that is composed of input-output pairs $(\vx, \vy)$ aggregated from all available datasets. Note that, among the options for operationalizing $f$, we follow the recent trend that uses LLMs with prompting, to harness their capabilities in understanding the longer and complex context (to jointly consider multiple samples from different datasets), which is not easily achievable by the traditional smaller models without performing additional labeling for and excessive training on them. Yet, the different challenge lies not only in the limitation that not all the external data samples can be accommodated within the context length of LLMs, but also in the fact that many of these samples may not be pertinent for generating valuable augmentations for $\mathcal{D}$. Therefore, addressing these critical issues necessitates answering the question: How can we selectively integrate only the pertinent instances from the extensive data store $\mathcal{C}$?

\subsubsection{Retrieving Relevant Instances}

We now turn to answer the question of retrieving contextually relevant instances from the data store $\mathcal{C}$, which is critical as it ensures that the data produced by LLMs is not only diverse and high-quality but also contextually coherent and aligned with the nuances of the target dataset $\mathcal{D}$. In the following, we first provide the general formulation of the retrieval and then propose our two specific instantiations of the retrieval for data augmentation. 

Formally, for a given input instance $\vq$, the goal of a retriever is to identify and fetch a ranked list of $k$ entries from a large corpus, which are deemed most relevant to the input, represented as follows: $\left\{ \vc_i \right\}_{i=1}^{k} = \texttt{Retriever}(\vq, \mathcal{C})$ where $\vc_i \in \mathcal{C}$. Here, $\vq$ can be a textual query, a set of keywords, or even a more complex structure depending on the application; the corpus $\mathcal{C}$ represents the entire database from which information is to be retrieved, which is typically a large collection of documents or passages; $\texttt{Retriever}$ is designed with keyword-based search algorithms~\cite{BM25} or neural embedding-based models~\cite{dpr}.

It is worth noting that, unlike typical retrieval approaches that primarily focus on sourcing relevant documents that are likely to contain the answers to the given query, in the context of our retrieval-augmented data augmentation scenario, we aim at fetching the relevant instances from other datasets, which are used as a source for generating the data along with the original samples. Therefore, these retrieved instances should ideally facilitate the generation of new and enriched samples. In addition, the instances to be retrieved can vary, which can be either complete input-output pairs or simply the inputs or outputs alone, depending on the specific requirements of data augmentation processes. We explain how we design retrieval in Section~\ref{method:augmentation}.

\begin{figure}[t!]
    \centering
    \includegraphics[width=0.975\linewidth]{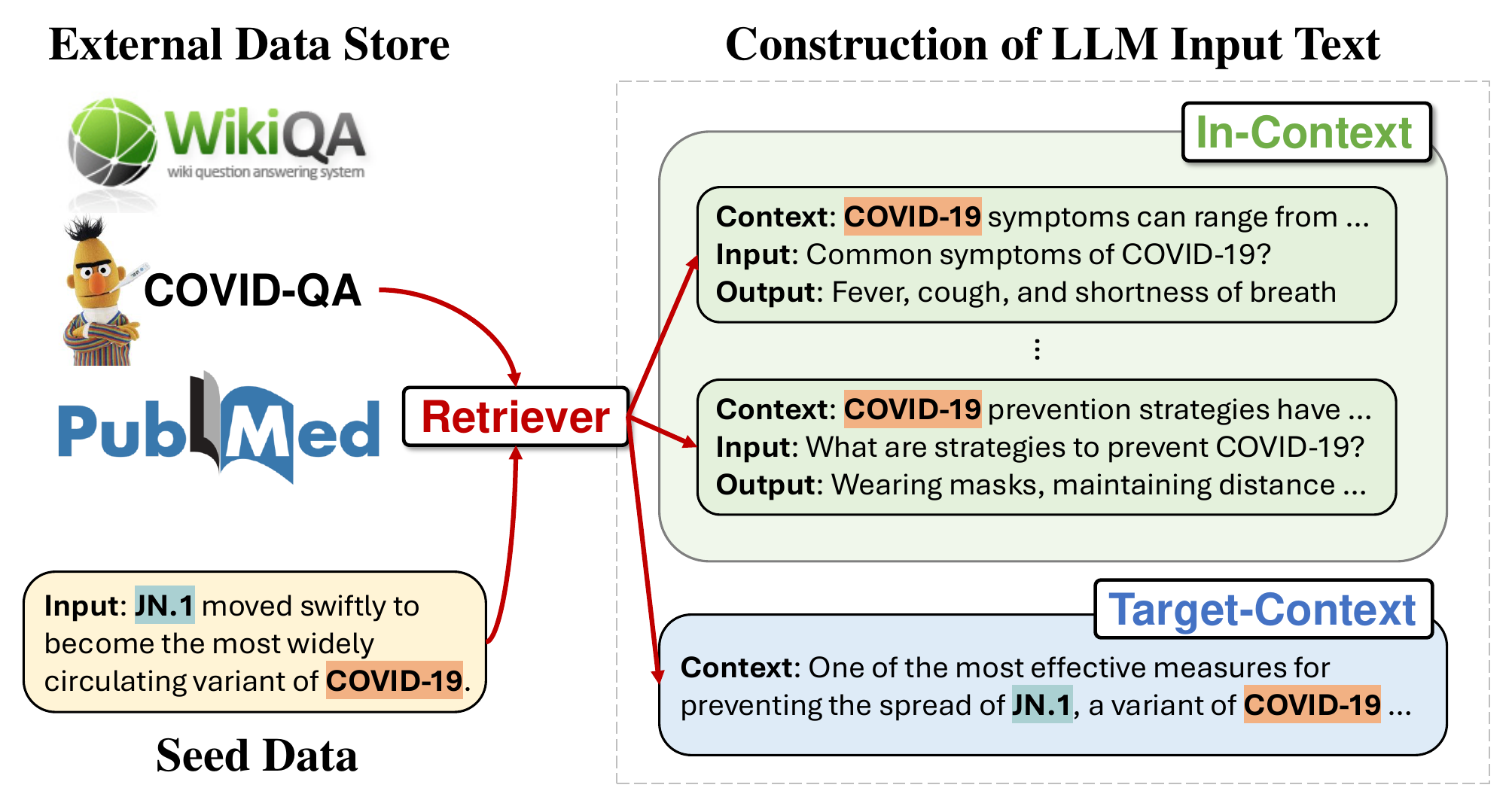}
    \vspace{-0.125in}
    \caption{\textbf{RADA Framework Overview}. We first retrieve the external instances (relevant to the seed data) from the external data store, and construct in-context and target-context of LLM prompts with the retrieved samples along with the seed data.}
    \label{fig:method}
    \vspace{-0.20in}
\end{figure}

\subsubsection{Retrieval for Data Augmentation}
\label{method:augmentation}

The input to LLMs can be viewed from two different perspectives: in-context learning which refers to their ability to learn from the input demonstrations; and task-solving where the model executes specific tasks requested by users (e.g., data augmentation). According to them, we propose two distinct instantiations of retrieval for LLM-powered data augmentation below (illustrated in Figure~\ref{fig:method}). 

\paragraph{Retrieval for In-Context Learning}

In-context learning plays a crucial role in enabling LLMs to align their outputs with the contextual cues provided in the input examples. Similarly, in the context of data augmentation, it may enable LLMs to learn from examples (e.g., input-output pairs) in the seed data, to generate new input-output pairs. However, in low-resource settings that we consider, the combination of data samples to provide as the examples in the input prompt is largely limited. This limitation highlights the advantage of our retrieval-augmented data augmentation framework, which can fill the input demonstrations with samples from external datasets. Yet, as not all the samples are relevant, we retrieve only the relevant samples based on the similarity between the sample in seed data $\mathcal{D}$ and the external sample in data store $\mathcal{C}$, as follows: $\left\{ \vc_i \right\}_{i=1}^{k} = \texttt{Retriever}(\vq, \mathcal{C})$ where $\vq \in \mathcal{D}$\footnote{The similarity calculation mechanism can vary, and, in this work, we consider the similarity between input queries.}. Mathematically, the combination of demonstrations to use as the LLM input is expanded to $O((k \times |\mathcal{D}|)^3)$ from $O(|\mathcal{D}|^3)$, where $|\mathcal{D}|$ is typically small in the low-resource setting and we assume using 3 demonstrations with top-$k$ sample retrievals.

\paragraph{Retrieval for Target Sample Generation}

Unlike in-context examples providing background information for data augmentation, the context to be retrieved and used here has a different goal, which should serve as a source for generating a complete input-output pair or one among them when given the other, depending on the specific use cases. Specifically, for question answering, a certain document can be used as a context to derive a query-answer pair along with their in-context examples. Another example is to provide a question as a context and then generate its answers, or vice versa to augment queries. It is worth noting that, while the usage of instances from the store $\mathcal{C}$ is different, their retrieval mechanism is the same as how we retrieve instances for in-context examples. Formally, $\left\{ \vc_i \right\}_{i=1}^{k} = \texttt{Retriever}(\vq, \mathcal{C})$ where $\vq$ can be, for question answering, either the document or the question from $\mathcal{D}$. Also, the augmented samples generated directly from the retrieved instances are similar in nature to the original samples, as we consider only the relevant top-$k$ instances for data augmentation, which can ensure a high degree of contextual coherence with seed samples, while being more diverse against the generation with seed.

\section{Experimental Setups}

In this section, we outline the experimental setups, including the datasets, models, and implementation details. We provide more details in Appendix~\ref{appendix:setups}.

\begin{table*}[t!]
\vspace{-0.1in}
\caption{\textbf{Training data augmentation results} on Covid QA, Policy QA, and Tech QA datasets with T5-base as the base model for training. In the second row, 10, 30, and 100 denote the number of initial seed data. We emphasize the best results in bold.}
\vspace{-0.1in}
\label{tab:main}
\small
\centering
\resizebox{\textwidth}{!}{
\renewcommand{\arraystretch}{0.9}
\begin{tabular}{lcccccccccccc}
\toprule

 & \multicolumn{3}{c}{\bf Covid QA} & \multicolumn{3}{c}{\bf Policy QA } & \multicolumn{3}{c}{\bf Tech QA } & \multicolumn{3}{c}{\bf Average } \\
\cmidrule(l{2pt}r{2pt}){2-4} \cmidrule(l{2pt}r{2pt}){5-7} \cmidrule(l{2pt}r{2pt}){8-10} \cmidrule(l{2pt}r{2pt}){11-13} 
\textbf{Methods} & 10 & 30 & 100 & 10 & 30 & 100 & 10 & 30 & 100 & 10 & 30 & 100\\

\midrule
\midrule


Seed Data & 53.94 & 66.50 & 68.44 & 7.62 & 20.20 & 27.79 & 9.54 & 17.91 & 36.52 & 23.70 & 34.87 & 44.25\\
\noalign{\vskip 0.25ex}\cdashline{1-13}\noalign{\vskip 0.75ex}

Augment w/ Seed Data & 61.15 & 64.70 & 65.06 & 28.04 &  27.20 & 25.96 &  39.17 & 40.45 & 41.27 & 42.79 & 44.11 & 44.10 \\

Self-Instruct & 61.85 & 61.89 & 64.38 &  26.92 & 27.54 & 27.27 & 32.68 & 39.11 & 37.77 & 40.48 & 42.85 & 43.14 \\

QA Generation & 52.75 & 51.03 & 39.10 & 19.33 & 20.63 & 21.13 & 29.98 & 31.06 & 32.21 & 34.02 & 34.24 & 30.81 \\

CQA Generation & 59.88 & 59.67 & 58.98  & 20.03 &21.78 & 20.34 & 21.82 & 20.24 & 23.42 & 33.91 & 33.90 & 34.25\\

Seed + External Data & 62.76 & 62.64 & 64.12 & 25.64 & 24.40 & \textbf{29.20} & 34.63 & 36.40 & 37.04 &41.01 & 41.15 & 43.45 \\

\noalign{\vskip 0.25ex}\cdashline{1-13}\noalign{\vskip 0.75ex}

\textbf{RADA (Ours)} & \textbf{67.49} & \textbf{68.15} & \textbf{68.57} & \textbf{29.23} & \textbf{28.49} & 29.18 & \textbf{40.81} & \textbf{44.37} & \textbf{46.93} & \textbf{45.84} & \textbf{47.00} & \textbf{48.23} \\

\bottomrule

\end{tabular}
}
\vspace{-0.15in}
\end{table*}

\subsection{Tasks and Datasets}

We validate our RADA on training data augmentation and test-time data augmentation scenarios. 

\paragraph{Training Data Augmentation}
The goal of training data augmentation is to expand the given samples, which is useful when new events occur that the model needs to adapt to, while having only limited data available for training. To test RADA with this scenario, we use three low-resource domain-specific datasets: Covid QA~\cite{covidqa} that is annotated by medical doctors for tackling the COVID-19 pandemic; Policy QA~\cite{policyqa} that is designed with specialized policies about website privacy; and Tech QA~\cite{techqa} that is constructed with questions on technical public forums for the IT domain. In addition, to simulate the low-resource settings, we assume 10, 30, and 100 instances are available for training, which are sampled from the training dataset. 

\vspace{-0.025in}
\paragraph{Test-Time Data Augmentation}
The assumption of test-time data augmentation is, on the other hand, more challenging, considering the situation where there is no data available for training due to strict privacy concerns (e.g., users or institutions may not want to share their own private data to train models). For this scenario, we select and use three specific domains from the MMLU dataset~\cite{mmlu} as it does not have direct training instances (aligned with our validation purpose), as well as using previous Covid QA, Policy QA, and Tech QA with no training samples available for this setup.

\vspace{-0.025in}
\paragraph{External Resources for Retrieval}
We construct the external data store serving as a retrieval source by aggregating samples from other datasets. Specifically, for Covid QA, Policy QA, and Tech QA designed for open-domain Question Answering (QA), we use Natural Questions (NQ)~\cite{nq} and labeled subset~\cite{domainspecificqa} of MS MARCO~\cite{msmarco}, covering broad domains with questions asked on web search. For MMLU that targets multi-choice QA, we use its official auxiliary data collected from similar datasets.

\subsection{Baselines and Our Model}

We compare our approach to several baselines including LLM-powered data augmentation methods.

\vspace{-0.05in}
\paragraph{Seed Data} 
It uses only the seed data for training models without extra data augmentation steps.

\vspace{-0.05in}
\paragraph{Augment w/ Seed Data}
It expands the seed data by generating new data instances from the seed data samples, where samples for in-context learning and target-context selection are randomly picked.

\vspace{-0.05in}
\paragraph{Self-Instruct}
It~\cite{self_instruct} aims to bootstrap new tasks only with limited seed examples, by incorporating the generated data instances in the data pool and leveraging them along with the seed data iteratively, where the samples in the pool are used to construct the in-context and target samples.

\vspace{-0.05in}
\paragraph{CQA Generation}
It~\cite{can_mrc_low} generates a context and then, based on it, subsequently generates a question-answer pair, where existing seed data samples are used for in-context learning. Its variant (\textbf{QA Generation}) simply generates a question-answer pair with in-context learning.

\vspace{-0.05in}
\paragraph{Seed + External Data}
It trains the models with the seed data instances as well as all the instances available in the external data pool. 

\vspace{-0.05in}
\paragraph{RADA}
This is our model that generates new data instances by retrieving samples (relevant to the seed data) from the external corpus and using them for in-context learning and target-context selection. 

\vspace{0.05in}
\noindent
We note that, for the test-time data augmentation scenario, since the samples having complete input-output pairs are unavailable, we cannot compare against the baselines requiring in-context examples; yet, RADA can run with only the target context.

\begin{table}[t!]
\caption{\textbf{Test-time data augmentation results} on subdomains of MMLU and domain-specific QA datasets. We use Llama2-7B as the base model for MMLU and T5-base for others.}
\vspace{-0.1in}
\label{tab:main_test}
\small
\centering
\resizebox{0.475\textwidth}{!}{
\renewcommand{\arraystretch}{0.95}
\begin{tabular}{lcccc}
\toprule
 \textbf{MMLU} & {\bf CS } &{\bf Biology } & {\bf Law } & {\bf Average } \\
 \midrule
 5-Shots w/ Training & 32.00 & 47.74 & 64.46 & 48.07 \\
 External Data & 48.00 & 54.52 & 66.12 & 56.21 \\

 \noalign{\vskip 0.25ex}\cdashline{1-5}\noalign{\vskip 0.75ex}

 {\bf RADA (Ours)} & {\bf 49.00} & {\bf 55.48} & {\bf 70.25} & {\bf 58.24} \\
 
\midrule
\midrule

 \textbf{Domain-Specific QA} & {\bf Covid } &{\bf Policy } & {\bf Tech } & {\bf Average } \\

\midrule


External Data & 53.54 & 19.40 & 13.46 & 28.80 \\

\noalign{\vskip 0.25ex}\cdashline{1-5}\noalign{\vskip 0.75ex}

\textbf{RADA (Ours)} &  \textbf{65.89} & \textbf{29.24} & \textbf{29.97} & \textbf{41.70}  \\

\bottomrule

\end{tabular}
}
\vspace{-0.125in}
\end{table}

\begin{figure*}[t!]
    \begin{minipage}{0.675\textwidth}
    \centering
    \includegraphics[width=0.975\linewidth]{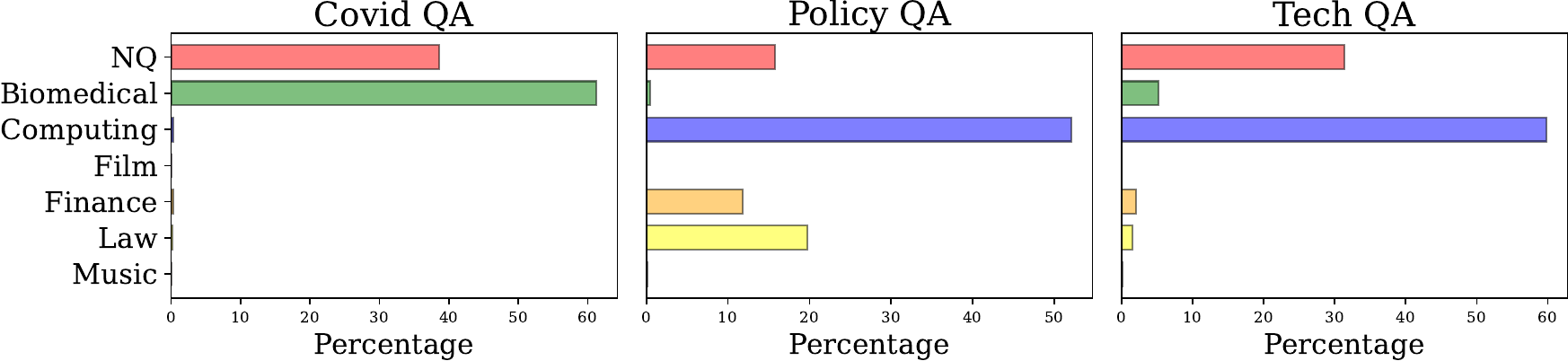}
    \vspace{-0.1in}
    \caption{\textbf{Breakdown results of retrieved instances} on three domain-specific QA datasets, where samples in the retrieval pool are one of Biomedical, Computing, Film, Finance, Law, and Music domains, as well as NQ (which covers general domains).}
    \label{fig:retrieve_result}
    \end{minipage}
    \hfill
    \begin{minipage}{0.30\textwidth}
    \small
    \centering
    \resizebox{0.975\textwidth}{!}{
    \renewcommand{\arraystretch}{1.05}
    \begin{tabular}{lcc}
    \toprule
    
    Domains & \textbf{Covid QA} & \textbf{Tech QA} \\
    
    \midrule
    \midrule
    
    All & 67.49 & 40.81 \\
    \noalign{\vskip 0.25ex}\cdashline{1-3}\noalign{\vskip 0.75ex}
    
    Biomedical & \textbf{67.75} & 40.09 \\
    Computing & 66.70 & \textbf{42.67} \\
    
    \bottomrule
    
    \end{tabular}
    }
    \vspace{-0.075in}
    \captionof{table}{\textbf{Results of the hand-crafted data store}, selectively using only the most suitable external domain as the retrieval pool for domain-specific QA.}
    \vspace{-0.125in}
    \label{tab:sensitivity}
    \end{minipage}
    \vspace{-0.05in}
\end{figure*}

\begin{figure*}[t!]
    \begin{minipage}{0.475\textwidth}
    \centering
    \includegraphics[width=0.975\linewidth]{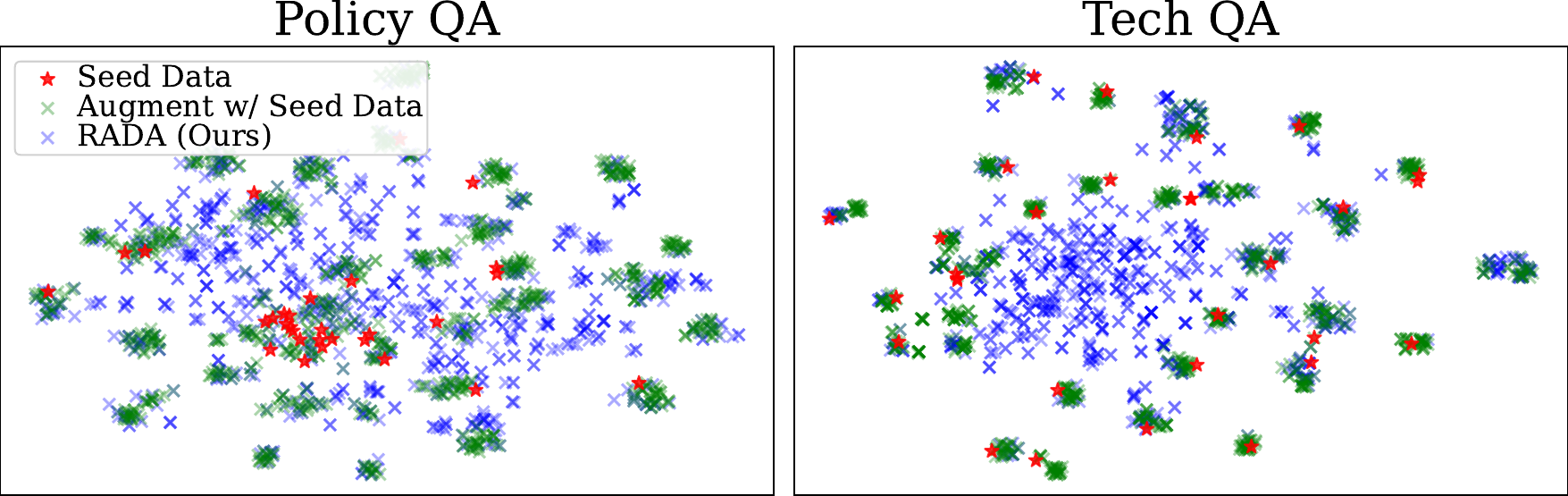}
    \vspace{-0.1in}
    \caption{\textbf{Embedding-space visualization results of samples} including the seed data and augmented data, with t-SNE. }
    \label{fig:tsne}
    \end{minipage}
    \hfill
    \begin{minipage}{0.475\textwidth}
    \small
    \centering
    \includegraphics[width=0.975\linewidth]{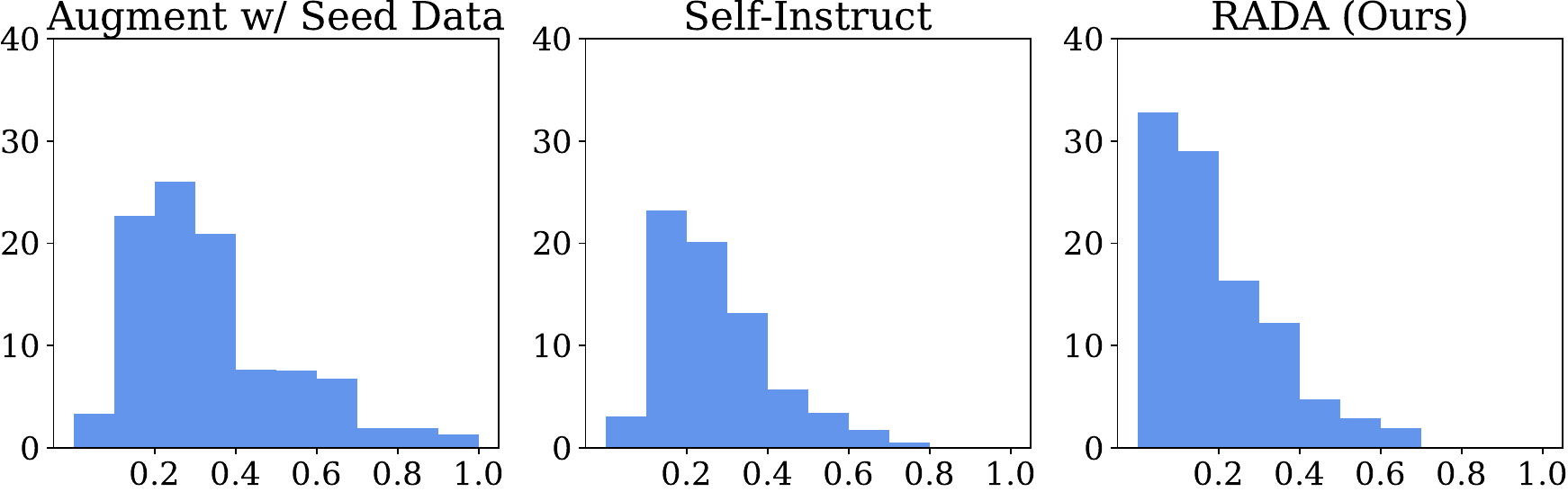}
    \vspace{-0.1in}
    \caption{\textbf{Results of ROUGE-L score distributions} measured between the seed data and generated data on Tech QA.}
    \label{fig:rouge}
    \end{minipage}
    \vspace{-0.15in}
\end{figure*}

\subsection{Implementation Details}

We use Llama2-7B-Chat~\cite{Llama2} as the basis for data augmentation across all methods. For fine-tuning we use either T5-base~\citep{t5} or Llama2-7B, to measure the effectiveness of different data augmentation approaches directly and to avoid data contamination issues as they are not trained on any downstream tasks/datasets. For the number of data augmented, unless otherwise stated, we produce samples amounting to 30 times that of the seed data and train models with the seed and generated data. A retriever used to retrieve instances is DistilBert TAS-B~\cite{ms-retriever}. We report results with the F1 score for Covid QA, Policy QA, and Tech QA datasets, and the accuracy for MMLU, following standard evaluation protocols. We provide prompts used to elicit data augmentation and answer generation in Appendix~\ref{appendix:setups}.

\section{Experimental Results}
\label{sec:exp}

\vspace{-0.05in}
\paragraph{Main Results}
We conduct experiments on two different data augmentation scenarios and report the results of training data augmentation in Table~\ref{tab:main}\footnote{We observe that the performance of Llama2 even after fine-tuning on the seed data and the augmented data is much inferior to T5-base on domain-specific QA; thus, we report results for them with T5 and further discuss it in Appendix~\ref{appendix:results}.} and the test-time augmentation results in Table~\ref{tab:main_test}. As shown in both tables, our RADA substantially outperforms all baselines across different settings (except for only one with Policy QA), demonstrating the effectiveness of our approach. In addition, one particular superior point of the Seed + External Data on Policy QA is not an unexpected result, since the number of initial seed data (100) is already large which is further coupled with a large number of external data samples (117,580), which may provide sufficient information to handle the task and whose number is actually much larger than the data used for RADA (30,100). Furthermore, as shown in Table~\ref{tab:main_test}, RADA is highly effective in the challenging test-time data augmentation scenario (where any data is unavailable for training), outperforming the model trained with all the external data instances. This may be attributed to our retrieval strategy for data augmentation, which results in generating samples that are relevant to the test data.

\begin{figure*}[!t]
    \begin{minipage}{0.675\linewidth}
        \centering
        \vspace{-0.03in}
        \includegraphics[width=1\linewidth]{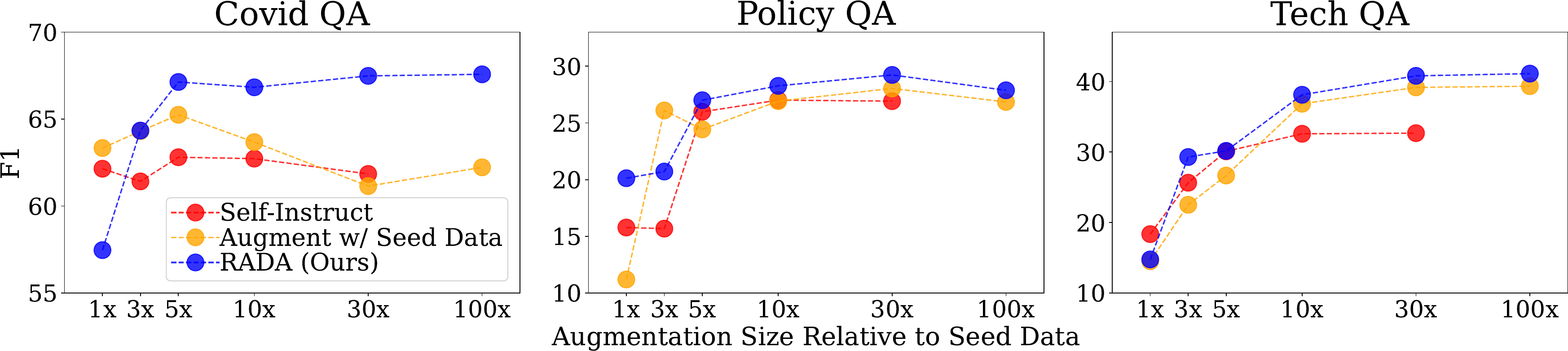} 
        \vspace{-0.3in}
        \caption{\textbf{Results of varying the augmentation size} on domain-specific QA, where we increase the size by factors of 1, 3, 5, 10, 30, and 100 relative to the seed data size.}
        \label{fig:ratio}
    \end{minipage}
    \hfill
    \begin{minipage}{0.31\linewidth}
        \resizebox{\linewidth}{!}{
            \renewcommand{\arraystretch}{1.05}            
     
            \begin{tabular}{lc}
            \toprule
            
            \textbf{Methods} & \textbf{Tech QA}  \\
            \midrule
            \midrule
            
            \textbf{RADA (Ours)} & \textbf{44.37} \\
            
            w/o In-context Retriever & 41.24  \\
            w/o Target-context Retriever & 34.42 \\
            w/o All Retrievers & 30.38  \\
            
            \bottomrule
            
            \end{tabular}
        }
        \vspace{-0.07in}
        \captionof{table}{\textbf{Ablation study} of the proposed RADA on the Tech QA dataset.}
        \vspace{-0.02in}
        \label{tab:ablation}
    \end{minipage}
    \vspace{-0.15in}
\end{figure*}

\paragraph{Analysis of Retrieval}
To understand which data instances are retrieved for data augmentation and what are their effectiveness, we conduct a comprehensive analysis. Firstly, we visualize the categories of retrieved instances for domain-specific QA in Figure~\ref{fig:retrieve_result}, which shows that (mostly) only the relevant instances are retrieved and used for data augmentation for each specific QA. For example, the Biomedical domain is the dominant field of retrieval source for Covid QA; meanwhile, the Computing domain is for Tech QA. In addition, to see the contribution of relevant retrieval, we restrict the retrieval domain to the one that is the most relevant to the given specific dataset. For example, we use only the Biomedical domain for Covid QA and the Computing domain for Tech QA. As shown in Table~\ref{tab:sensitivity}, we observe that when manipulating the retrieval pool, the performance further increases (as instances from irrelevant domains are not retrieved), which reaffirms the effectiveness of retrieval and its room for improvement for data augmentation.

\paragraph{Analysis of Augmented Data Diversity}

A notable advantage of RADA is that it intuitively can generate more diverse samples than what could be achieved by existing data augmentation approaches that use the seed data alone, by augmenting this process with the retrieval from external data samples. To measure this ability, we visualize the embedding space of the augmented samples across different models in Figure~\ref{fig:tsne} and report their lexical overlaps in Figure~\ref{fig:rouge}. Specifically, for the visualization, we first embed the generated instances with Sentence-BERT~\cite{reimers-2019-sentence-bert} into the latent space and project them with t-SNE~\cite{tsne}. From this, we observe that, unlike Augment w/ Seed Data whose generated samples are close to the seed data, the samples generated from RADA are broadly dispersed across the space. Further, we measure the max ROUGE-L scores between the seed instances and the generated instances where lower scores indicate higher diversity. As shown in Figure~\ref{fig:rouge}, RADA generates distinct samples to the seed data thanks to retrieving and utilizing the external contexts beyond the seed data, unlike baselines that rely solely on it.

\paragraph{Analysis of Augmented Data Size}

To see how the performance changes as a function of the size of augmented data samples, we vary the augmentation size relative to the seed data size by a factor of 1, 3, 5, and up to 100 times and report the results in Figure~\ref{fig:ratio}\footnote{Due to the cost of running Self-Instruct, we are not able to generate its samples for the 100 times augmentation-level.}. Firstly, when the amount of augmented data is very small, baseline performances are comparable with RADA since the data samples that can be generated from the seed data alone can have a certain diversity level as we augment only a small amount. However, as the size of augmentation expands, RADA consistently outperforms baselines, showcasing its ability to generate broader and richer samples through retrieval augmentation, while the performance starts to converge after a 100-time increase in data augmentation.

\paragraph{Ablation Study}
To see how each component of RADA affects the overall performance, we conduct an ablation study where we replace our in-context and target-context retrieval modules with random retrievals. As shown in Table~\ref{tab:ablation}, we observe that, without retrieving relevant instances, the performances drop substantially since irrelevant samples (to the target tasks/datasets) are used to construct the in-context examples and target context, leading to generating the samples not useful for them. Furthermore, the target-context retriever is particularly important for data augmentation, since this context is used to directly derive the instances for training.

\begin{table}[t!]
\caption{\textbf{Results of another LLM (ChatGPT)} for data augmentation on domain-specific QA with seed examples of 10.}
\vspace{-0.1in}
\label{tab:main_chatgpt}
\small
\centering
\resizebox{0.475\textwidth}{!}{
\renewcommand{\arraystretch}{1.0}
\begin{tabular}{lcccc}
\toprule

 & {\bf Covid } &{\bf Policy } & {\bf Tech } &{\bf Average } \\

\midrule
\midrule


Self-Instruct & 57.86 & 26.20 & 33.42 &  39.16 \\ 
CQA Generation & 65.64 &  27.20 & 34.16 & 42.33 \\

\noalign{\vskip 0.25ex}\cdashline{1-5}\noalign{\vskip 0.75ex}

\textbf{RADA (Ours)} &  \textbf{67.19} & \textbf{28.59} & \textbf{36.17} &  \textbf{43.98}  \\

\bottomrule

\end{tabular}
}
\vspace{-0.125in}
\end{table}

\paragraph{Analysis of Using Different LLMs}
Finally, we conduct an auxiliary analysis to see whether the superiority of RADA is consistent across different LLMs, compared to existing baselines. In particular, we use ChatGPT 3.5 (released on June 13, 2023) as the basis model for data augmentation, and report the results in Table~\ref{tab:main_chatgpt}. From this, we observe that RADA significantly outperforms baselines with another LLM, demonstrating its robustness across different LLMs for data augmentation.

\vspace{-0.05in}
\section{Conclusion}
\vspace{-0.075in}

In this work, we pointed out the limitation of existing data augmentation approaches that use the seed data alone for low-resource domain tasks, leading to generating suboptimal and less diverse instances, despite the existence of plenty of external samples available. Inspired by this, we proposed the LLM-powered Retrieval-Augmented Data Augmentation (RADA) framework, which augments the seed data by leveraging the samples retrieved from the external data store based on their relevance with the seed data, during data augmentation. Specifically, the input to LLMs for data augmentation can be viewed from two different angles of in-context examples and task-solving context, and we constructed them through samples from within and across the seed data and the retrieved data. Through extensive evaluation results on multiple datasets with training and test-time data augmentation scenarios, we showed that RADA outperforms strong LLM-powered data augmentation baselines substantially. In addition, our findings reveal that the data samples generated from our approach are much more diverse against baselines while being relevant to the seed data, due to leveraging retrieval for data augmentation. We believe that RADA will pave the way for enhancing the model performances on realistic low-resource domain-specific tasks/datasets, which have arisen as very important problems recently due to the limited availability and privacy concerns of data.

\section*{Limitations}
In this section, we faithfully discuss some remaining room for improvements to our RADA framework. First of all, the effectiveness of our retrieval-augmentation approach (by its nature) depends on the quality and relevance of the external data store. Thus, the performance of RADA may degenerate if the retrieval source is not truly aligned with our seed data, and we leave exploring this new setting as future work. Also, investigating the scenario of continuously updating the retrieval pool over time would be interesting for future work as well. On the other hand, due to the heavy cost of fine-tuning LLMs, data sample efficiency (i.e., reducing the amount of samples to train while maintaining the model performance) becomes an important agenda. While we do have some preliminary results on filtering augmented samples in Appendix~\ref{appendix:results}, it would be interesting to developing more on this direction.

\section*{Ethics Statement}

While our RADA is superior in generating more diverse and high-quality samples (compared to existing data augmentation approaches), its performance is not flawless: the retriever might retrieve offensive or harmful instances for data augmentation, and the generator might produce plausible yet factually incorrect instances. Therefore, it may be carefully used for mission-critical domains, such as biomedical or legal fields, (perhaps with the help of domain-experts during the augmentation process).

\bibliography{custom}

\appendix

\clearpage

\section{Additional Experimental Setups}
\label{appendix:setups}

\paragraph{Fine-tuning Details}

We provide more details on how to fine-tune models on the seed and augmented data samples. Firstly, for T5-base, we train it over 5 epochs with a batch size of 8 and a learning rate of 3$\times 10^{-5}$, selecting the best epoch to report the performance with inference. For Llama-7B, to train it with our computational resources available, we use the QLORA~\citep{qlora} technique, on which we use the epoch size of 30, the batch size of 1, and the learning rate of 2$\times 10^{-4}$. Lastly, we report the fine-tuning results with a single run.

\paragraph{Prompts}

The prompt used to elicit the data augmentation is provided in Table~\ref{tab:prompt}. For the domain-specific datasets including Covid QA, Policy QA, and Tech QA, we use the following prompt to generate the answer: "Context: \{ \} Question: \{ \} Answer: ". For the MMLU dataset, we use the following prompt: "Question: \{ \} Answer Options: \{ \} Answer:" where 5-shot examples prepended are the same as the one in the official code repository\footnote{https://github.com/hendrycks/test}.

\paragraph{Computational Resources and Time}
We train and inference all baselines and our model by using one of the TITAN RTX, NVIDIA GeForce RTX 3080, NVIDIA GeForce RTX 3090, NVIDIA RTX A4000, NVIDIA RTX A5000, and Quadro RTX 8000 GPUs, depending on their availability at the time of run. The time required for training RADA ranges from a few minutes to about one and half day, which also depends on the number of the augmented data used for model fine-tuning.

\paragraph{Deep Learning Libraries}

In our experiments, we utilize the deep learning libraries as follows: PyTorch~\citep{pytorch}, Transformers~\citep{hf_transformers}, SentenceTransformers~\citep{sentencebert}, and BEIR~\citep{beir}. We will release the specific requirements for reproducing our results, upon releasing the code.

\begin{table}[t!]
\caption{Training time results on Covid QA, where we use T5 or Llama as the base for fine-tuning on augmented data.}
\vspace{-0.1in}
\label{tab:llama}
\small
\centering
\resizebox{0.475\textwidth}{!}{
\renewcommand{\arraystretch}{1.0}
\begin{tabular}{llcccc}
\toprule

{\bf \# of seed} & {\bf Bases} & {\bf 0-shot} &{\bf 5-shot} & {\bf Seed} &{\bf RADA (Ours)} \\

\midrule
\midrule


\multirowcell{2}[-0.1ex][l]{10} 
& T5 & N/A & N/A & \textbf{53.94} & \textbf{67.49} \\ 
& Llama2 & 12.79 & 16.43 & 50.62 & 56.50 \\ 

\midrule

\multirowcell{2}[-0.1ex][l]{30} 
& T5 & N/A & N/A & \textbf{66.50} & \textbf{68.15} \\
& Llama2 & 12.79 & 16.43 & 55.48 & 53.62 \\

\bottomrule

\end{tabular}
}
\end{table}
\begin{figure}
    \centering
    \includegraphics[width=0.975\linewidth]{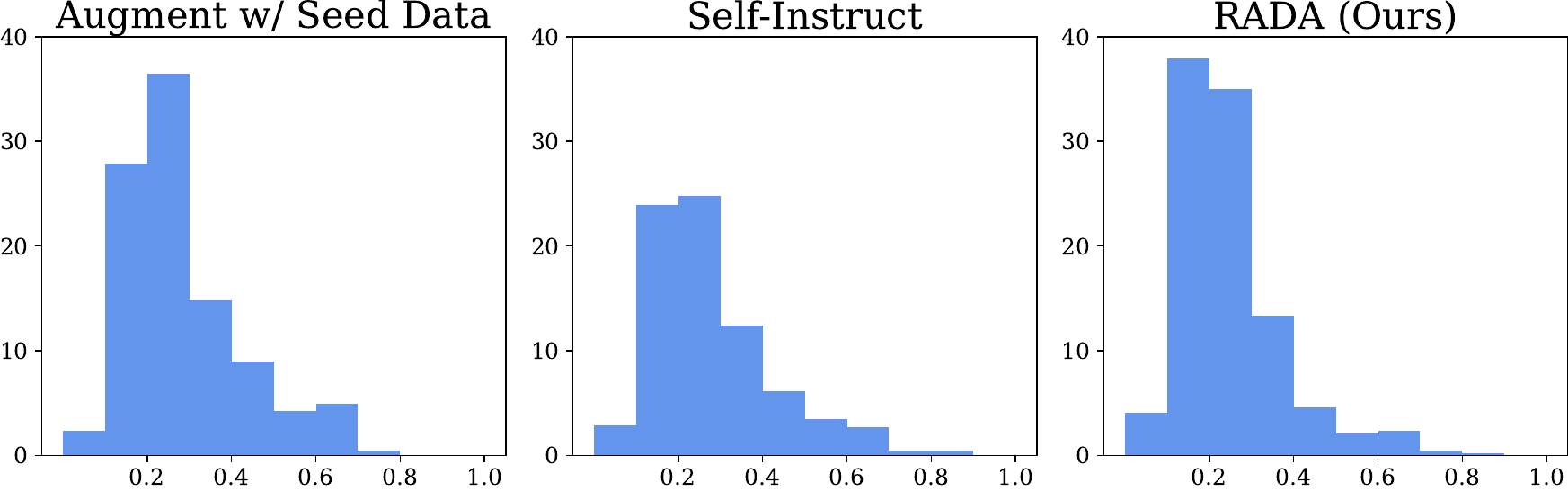}
    \vspace{-0.1in}
    \caption{\textbf{Results of ROUGE-L score distributions} measured between the seed data and generated data on Covid QA.}
    \label{fig:rouge:covid}
\end{figure}
\begin{figure}
    \centering
    \includegraphics[width=0.975\linewidth]{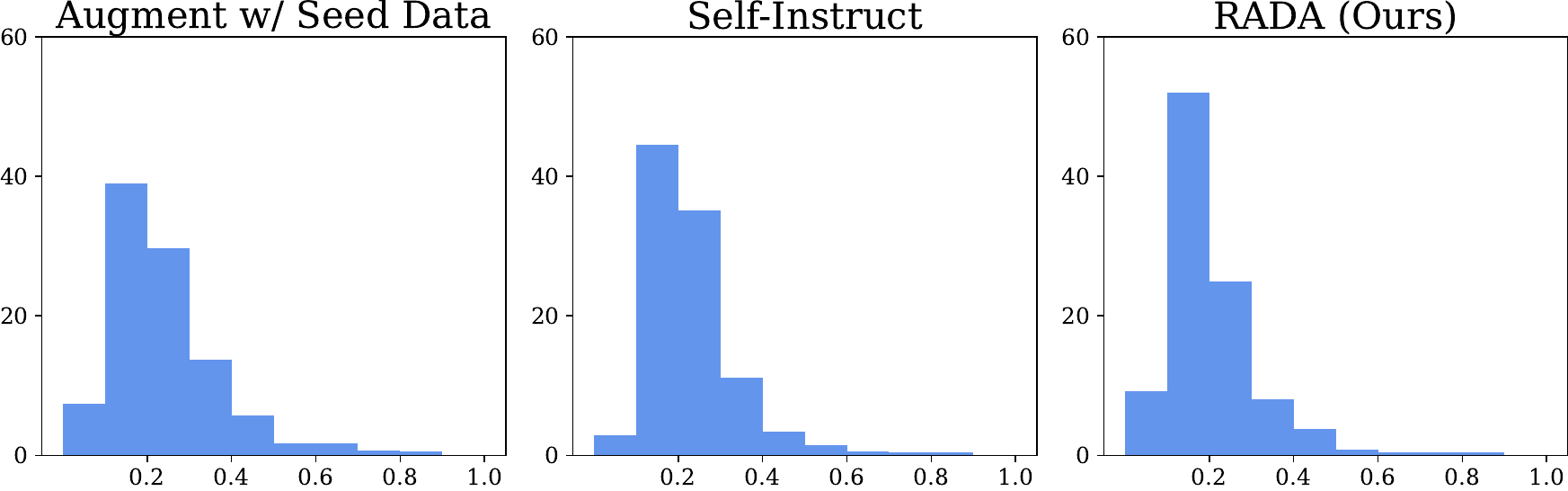}
    \vspace{-0.1in}
    \caption{\textbf{Results of ROUGE-L score distributions} measured between the seed data and generated data on Policy QA.}
    \label{fig:rouge:policy}
\end{figure}
 
\begin{table*}[t!]
\caption{\textbf{Results of various filtering mechanisms} on domain-specific QA datasets with training data augmentation settings.}
\vspace{-0.1in}
\label{tab:filtering}
\small
\centering
\resizebox{\textwidth}{!}{
\renewcommand{\arraystretch}{1.0}
\begin{tabular}{lcccccccccccc}
\toprule

 & \multicolumn{3}{c}{\bf Covid QA} & \multicolumn{3}{c}{\bf Policy QA } & \multicolumn{3}{c}{\bf Tech QA } & \multicolumn{3}{c}{\bf Average } \\
\cmidrule(l{2pt}r{2pt}){2-4} \cmidrule(l{2pt}r{2pt}){5-7} \cmidrule(l{2pt}r{2pt}){8-10} \cmidrule(l{2pt}r{2pt}){11-13} 
\textbf{Methods} & 10 & 30 & 100 & 10 & 30 & 100 & 10 & 30 & 100 & 10 & 30 & 100\\

\midrule
\midrule


\textbf{RADA (Ours)} & \textbf{67.49} & \textbf{68.15} & \textbf{68.57} & \textbf{29.23} & \textbf{28.49} & {\bf 29.18} & \textbf{40.81} & 44.37 & \textbf{46.93} & \textbf{45.84} & \textbf{47.00} & \textbf{48.23} \\

\noalign{\vskip 0.25ex}\cdashline{1-13}\noalign{\vskip 0.75ex}

w/ ROUGE-based Filtering & 66.21 & 67.25 & 66.84 & 28.35 & 28.09 & 28.31 & 37.75 & {\bf 44.64} & 46.74 & 44.10 & 46.66 & 47.30 \\

w/ Embedding-based Filtering & 67.19 & 67.67 & 67.27 & 28.62 & 28.13 & 28.65 & 40.02 & {\bf 44.64} & 46.74 & 45.27 & 46.82 & 47.55  \\

\noalign{\vskip 0.25ex}\cdashline{1-13}\noalign{\vskip 0.75ex}

w/o Answer Filtering & 66.78 & 66.65 & 67.09 & 28.78 & 28.44 & 29.12 & 40.55 & 42.43 & 42.56 & 45.37 & 45.84 & 46.26  \\



\bottomrule

\end{tabular}
}
\vspace{-0.0in}
\end{table*}

\section{Additional Experimental Results}
\label{appendix:results}

\paragraph{More Analysis on Data Diversity}
In addition to the result of ROUGE-L score distributions on Tech QA in Figure~\ref{fig:rouge}, we provide results on Covid QA and Policy QA in Figure~\ref{fig:rouge:covid} and Figure~\ref{fig:rouge:policy}, respectively. From this, we consistently observe that the proposed RADA generates diverse instances during data augmentation, compared to other baselines.

\paragraph{Results of Llama on Domain-Specific QA}

Here we discuss the training data augmentation results of Llama on domain-specific QA data (such as Covid QA). Specifically, in Table~\ref{tab:llama}, we report its 0-shot and 5-shot performances, as well as its fine-tuning performances on seed data and augmented data. As shown in Table~\ref{tab:llama}, despite the large number of parameters that Llama2-7B has (which is ten times larger than T5), we observe that Llama2 is inferior to T5. We conjecture that this may be because the general massive corpus used to pre-train Llama2 has little (to no) overlap or relevance with instances in domain-specific tasks. In other words, eliciting the domain-specific ability of Llama2 with fine-tuning may be largely suboptimal, when it does not have internalized knowledge about its corresponding domain-specific tasks. In addition, this result may further highlight the fact that not all the larger models perform always better than the smaller models in low-resource settings, which gives us a promise to take advantage of computational efficiency, especially when dealing with extreme domain-specific tasks, or that specific LLMs may be required to handle each specific domain.

\paragraph{Results with Filtering}
We try various filtering approaches on the augmented data to fine-tune models with only the samples of high quality. Specifically, to further promote diversity in the generated samples from our RADA, we filter samples if they are similar to the already generated samples, based on their ROUGE scores or their embedding-level distances. Then, as shown in Table~\ref{tab:filtering}, these filtering techniques do not improve the model performance. This may further strengthen our claim that the augmented instances from RADA are already very diverse but also relevant to the seed data, which does not necessitate additional filtering mechanisms. On the other hand, if we relax the assumption that the passage should include the answer to the question for domain-specific QA, and subsequently do not apply the filtering strategy (checking the inclusiveness), the performance drops slightly in Table~\ref{tab:filtering}.

\paragraph{Quantitative Analysis}
In Table~\ref{tab:examples:covid},~\ref{tab:examples:policy},~\ref{tab:examples:tech}, we provide examples of the augmented instances across different methods on Covid QA, Policy QA, and Tech QA. A key finding from these results is that the existing approach that uses only the seed data results in a limited diversity of generated samples, unlike our RADA which generates distinct yet contextually coherent samples with the seed data, thanks to the retrieval of relevant external samples.

\begin{table*}[t!]
    \caption{A list of prompts that we use for data augmentation with the proposed RADA framework. It is worth noting that the variable inside the parentheses \{\} is replaced with its actual string (e.g., context, question, answer options, and answer). Also, the last sentence of the prompt represents the target context, which is used as the main source of information to generate the augmented instance. For MMLU, we use the combinations of Version 1 and Version 2 for data augmentation.}
    \label{tab:prompt}
    \vspace{-0.075in}
    \resizebox{1\textwidth}{!}{
        \begin{tabular}{ll}
        \toprule
        \multicolumn{1}{p{.15\textwidth}}{\textbf{Types}} & \makecell{\multicolumn{1}{p{.85\textwidth}}{\textbf{Prompts}}} \\
        \midrule
        \multicolumn{1}{p{.15\textwidth}}{\textbf{Domain-specific QA}} & 
        \makecell{
            \multicolumn{1}{p{.85\textwidth}}{I want you to act as a question and answer generator. 
Your goal is to create an extractive question-answer pair based on a given context.
The answer to the question must be a specific span from the given context.} \\
\multicolumn{1}{p{.85\textwidth}}{Context: \{context 1\}} \\
\multicolumn{1}{p{.85\textwidth}}{Question: \{question 1\}} \\
\multicolumn{1}{p{.85\textwidth}}{Answer: \{answer 1\}} \\
\multicolumn{1}{p{.85\textwidth}}{Context: \{context 2\}} \\
\multicolumn{1}{p{.85\textwidth}}{Question: \{question 2\}} \\
\multicolumn{1}{p{.85\textwidth}}{Answer: \{answer 2\}} \\
\multicolumn{1}{p{.85\textwidth}}{Context: \{context 3\}} \\
\multicolumn{1}{p{.85\textwidth}}{Question: \{question 3\}} \\
\multicolumn{1}{p{.85\textwidth}}{Answer: \{answer 3\}} \\
\multicolumn{1}{p{.85\textwidth}}{Context: \{context\}} \\
        }\\
        \midrule
        \multicolumn{1}{p{.15\textwidth}}{\textbf{MMLU \;\;\;\; (Version 1)}} & 
        \makecell{
            \multicolumn{1}{p{.85\textwidth}}{I want you to act as an answer options and answer generator. 
Your goal is to create four answer options and the answer pair based on a given question.
The answer must be one of the generated answer options.} \\
\multicolumn{1}{p{.85\textwidth}}{Question: \{question 1\}} \\
\multicolumn{1}{p{.85\textwidth}}{Answer Options: \{answer options 1\}} \\
\multicolumn{1}{p{.85\textwidth}}{Answer: \{answer 1\}} \\
\multicolumn{1}{p{.85\textwidth}}{Question: \{question 2\}} \\
\multicolumn{1}{p{.85\textwidth}}{Answer Options: \{answer options 2\}} \\
\multicolumn{1}{p{.85\textwidth}}{Answer: \{answer 2\}} \\
\multicolumn{1}{p{.85\textwidth}}{Question: \{question 3\}} \\
\multicolumn{1}{p{.85\textwidth}}{Answer Options: \{answer options 3\}} \\
\multicolumn{1}{p{.85\textwidth}}{Answer: \{answer 3\}} \\
\multicolumn{1}{p{.85\textwidth}}{Question: \{question\}}} \\
\noalign{\vskip 0.25ex}\cdashline{1-2}\noalign{\vskip 0.75ex}
         \multicolumn{1}{p{.15\textwidth}}{\textbf{MMLU \;\;\;\; (Version 2)}} &
        \makecell{
\multicolumn{1}{p{.85\textwidth}}{I want you to act as a question and answer generator. 
Your goal is to create an extractive question-answer pair based on the given answer options.
The answer to the question must be selected from the given answer options.} \\
\multicolumn{1}{p{.85\textwidth}}{Answer Options: \{answer options 1\}} \\
\multicolumn{1}{p{.85\textwidth}}{Question: \{question 1\}} \\
\multicolumn{1}{p{.85\textwidth}}{Answer: \{answer 1\}} \\
\multicolumn{1}{p{.85\textwidth}}{Answer Options: \{answer options 2\}} \\
\multicolumn{1}{p{.85\textwidth}}{Question: \{question 2\}} \\
\multicolumn{1}{p{.85\textwidth}}{Answer: \{answer 2\}} \\
\multicolumn{1}{p{.85\textwidth}}{Answer Options: \{answer options 3\}} \\
\multicolumn{1}{p{.85\textwidth}}{Question: \{question 3\}} \\
\multicolumn{1}{p{.85\textwidth}}{Answer: \{answer 3\}} \\
\multicolumn{1}{p{.85\textwidth}}{Answer Options: \{answer options\}} 
}\\
        \bottomrule
        \end{tabular}
    }
\end{table*}

\begin{table*}[t!]
    \caption{The example question-answer pairs generated from different models on Covid QA.}
    \label{tab:examples:covid}
    \vspace{-0.1in}
    \resizebox{1\textwidth}{!}{
        \begin{tabular}{ll}
        \toprule
        \multicolumn{1}{p{.15\textwidth}}{\textbf{Types}} & \makecell{\multicolumn{1}{p{.85\textwidth}}{\textbf{Samples}}} \\
        \midrule
        \multicolumn{1}{p{.15\textwidth}}{\textbf{Augment \;\;\;\;\; w/ Seed Data}} & 
        \makecell{
            \multicolumn{1}{p{.85\textwidth}}{{\bf Context}: polymerase chain reaction testing, the time lag between hospitalization and reporting was longer for early cases compared with that of more recent cases. Among the seven locations reporting importation, the total volume of inbound passengers from China was m = 63.1 million per year in 2017 [9] , of which 100q = 2.1\% were from Wuhan [10] , a home of n = 19.0 million people as the catchment population of Wuhan airport. }\\
            \multicolumn{1}{p{.85\textwidth}}{{\bf Generated Question}: What was the total volume of inbound passengers from China to the seven locations reporting importation in 2017?} \\
            \multicolumn{1}{p{.85\textwidth}}{{\bf Generated Answer}: 63.1 million per year} 
        }\\
        \midrule
        \multicolumn{1}{p{.15\textwidth}}{\textbf{Self-Instruct}} & 
        \makecell{
            \multicolumn{1}{p{.85\textwidth}}{{\bf Context}: polymerase chain reaction testing, the time lag between hospitalization and reporting was longer for early cases compared with that of more recent cases. Among the seven locations reporting importation, the total volume of inbound passengers from China was m = 63.1 million per year in 2017 [9] , of which 100q = 2.1\% were from Wuhan [10] , a home of n = 19.0 million people as the catchment population of Wuhan airport. }\\
            \multicolumn{1}{p{.85\textwidth}}{{\bf Generated Question}: What is the total volume of inbound passengers from China?}  \\
            \multicolumn{1}{p{.85\textwidth}}{{\bf Generated Answer}: m = 63.1 million per year in 2017} 
        } \\
        \midrule
        \multicolumn{1}{p{.15\textwidth}}{\textbf{RADA (Ours)}} & 
        \makecell{
            \multicolumn{1}{p{.85\textwidth}}{{\bf Context}: MYTH: The flu is just a bad cold. Influenza may cause bad cold symptoms, like sore throat, runny nose, sneezing, hoarseness, and cough. But in the United States alone, 36,000 people die and more than 200,000 are hospitalized each year because of the flu. During the 2017/18 flu season, flu activity has significantly increased throughout the majority of the country with the A(H3N2) viruses predominating so far, according to the Centers for Disease Control (CDC).}\\
            \multicolumn{1}{p{.85\textwidth}}{{\bf Generated Question}: What are the consequences of getting the flu?} \\
            \multicolumn{1}{p{.85\textwidth}}{{\bf Generated Answer}: 36,000 people die and more than 200,000 are hospitalized each year because of the flu.} 
        } \\
        \bottomrule
        \end{tabular}
    }
\end{table*}
\begin{table*}[t!]
    \caption{The example question-answer pairs generated from different models on Policy QA.}
    \label{tab:examples:policy}
    \vspace{-0.1in}
    \resizebox{1\textwidth}{!}{
        \begin{tabular}{ll}
        \toprule
        \multicolumn{1}{p{.15\textwidth}}{\textbf{Types}} & \makecell{\multicolumn{1}{p{.85\textwidth}}{\textbf{Samples}}} \\
        \midrule
        \multicolumn{1}{p{.15\textwidth}}{\textbf{Augment \;\;\;\;\; w/ Seed Data}} & 
        \makecell{
            \multicolumn{1}{p{.85\textwidth}}{{\bf Context}: IP Addresses. We use your IP address to help diagnose problems with our server, and to administer our Web site. Your IP address is used to help identify you and, in the case of the Coffee Review Marketplace, your shopping cart and to gather broad demographic information in the interest of improving our site and service.}\\
            \multicolumn{1}{p{.85\textwidth}}{{\bf Generated Question}: How does the company use IP addresses?}  \\
            \multicolumn{1}{p{.85\textwidth}}{{\bf Generated Answer}: to help diagnose problems with our server, and to administer our Web site.} 
        }\\
        \midrule
        \multicolumn{1}{p{.15\textwidth}}{\textbf{Self-Instruct}} & 
        \makecell{
            \multicolumn{1}{p{.85\textwidth}}{{\bf Context}: IP Addresses. We use your IP address to help diagnose problems with our server, and to administer our Web site. Your IP address is used to help identify you and, in the case of the Coffee Review Marketplace, your shopping cart and to gather broad demographic information in the interest of improving our site and service.}\\
            \multicolumn{1}{p{.85\textwidth}}{{\bf Generated Question}: What information does the company use IP addresses for?} \\
            \multicolumn{1}{p{.85\textwidth}}{{\bf Generated Answer}: to help diagnose problems with our server, and to administer our Web site. Your IP address is used to help identify you and, in the case of the Coffee Review Marketplace, your shopping cart and to gather broad demographic information in the interest of improving our site and service.} 
        } \\
        \midrule
        \multicolumn{1}{p{.15\textwidth}}{\textbf{RADA (Ours)}} & 
        \makecell{
            \multicolumn{1}{p{.85\textwidth}}{{\bf Context}: The 1998 Data Protection Act sets the rules for how UK businesses can gather, store and use people confidential data. The Act is organised around eight principles of data protection. They include the idea that personal data should be obtained and processed fairly and lawfully, that only the right data should be used, that data should be accurate and so on. Principle Seven covers data security. It states:}\\
            \multicolumn{1}{p{.85\textwidth}}{{\bf Generated Question}: What is the main purpose of the 1998 Data Protection Act?}  \\ 
            \multicolumn{1}{p{.85\textwidth}}{{\bf Generated Answer}: The 1998 Data Protection Act sets the rules for how UK businesses can gather, store and use people confidential data.} 
        } \\
        \bottomrule
        \end{tabular}
    }
\end{table*}
\begin{table*}[t!]
    \caption{The example question-answer pairs generated from different models on Tech QA.}
    \label{tab:examples:tech}
    \vspace{-0.1in}
    \resizebox{1 \textwidth}{!}{
        \begin{tabular}{ll}
        \toprule
        \multicolumn{1}{p{.15\textwidth}}{\textbf{Types}} & \makecell{\multicolumn{1}{p{.85\textwidth}}{\textbf{Samples}}} \\
        \midrule
        \multicolumn{1}{p{.15\textwidth}}{\textbf{Augment \;\;\;\;\; w/ Seed Data}} & 
        \makecell{
            \multicolumn{1}{p{.85\textwidth}}{{\bf Context}: VULNERABILITY DETAILS}\\ \\
            \multicolumn{1}{p{.85\textwidth}}{CVEID: CVE-2015-1920 [http://cve.mitre.org/cgi-bin/cvename.cgi?name=CVE-2015-1920]}\\ 
            \multicolumn{1}{p{.85\textwidth}}{DESCRIPTION: WebSphere Application Server could allow a remote attacker to execute arbitrary code by connecting to a management port and executing a specific sequence of instructions.}\\ 
            \multicolumn{1}{p{.85\textwidth}}{CVSS Base Score: 9.3}\\ 
            \multicolumn{1}{p{.85\textwidth}}{CVSS Temporal Score: See https://exchange.xforce.ibmcloud.com/vulnerabilities/102404 [https://exchange.xforce.ibmcloud.com/vulnerabilities/102404] for the current score}\\ 
            \multicolumn{1}{p{.85\textwidth}}{CVSS Environmental Score*: Undefined}\\ 
            \multicolumn{1}{p{.85\textwidth}}{CVSS Vector: (AV:N/AC:M/Au:N/C:C/I:C/A:C)}\\
            \multicolumn{1}{p{.85\textwidth}}{{\bf Generated Question}: What is the CVSS Base Score of the vulnerability in IBM WebSphere Application Server?}  \\ 
            \multicolumn{1}{p{.85\textwidth}}{{\bf Generated Answer}: The CVSS Base Score of the vulnerability in IBM WebSphere Application Server is 9.3.} 
        }\\
        \midrule
        \multicolumn{1}{p{.15\textwidth}}{\textbf{Self-Instruct}} & 
        \makecell{
            \multicolumn{1}{p{.85\textwidth}}{{\bf Context}: VULNERABILITY DETAILS}\\ \\
            \multicolumn{1}{p{.85\textwidth}}{CVEID: CVE-2015-1920 [http://cve.mitre.org/cgi-bin/cvename.cgi?name=CVE-2015-1920]}\\ 
            \multicolumn{1}{p{.85\textwidth}}{DESCRIPTION: WebSphere Application Server could allow a remote attacker to execute arbitrary code by connecting to a management port and executing a specific sequence of instructions.}\\ 
            \multicolumn{1}{p{.85\textwidth}}{CVSS Base Score: 9.3}\\ 
            \multicolumn{1}{p{.85\textwidth}}{CVSS Temporal Score: See https://exchange.xforce.ibmcloud.com/vulnerabilities/102404 [https://exchange.xforce.ibmcloud.com/vulnerabilities/102404] for the current score}\\ 
            \multicolumn{1}{p{.85\textwidth}}{CVSS Environmental Score*: Undefined}\\ 
            \multicolumn{1}{p{.85\textwidth}}{CVSS Vector: (AV:N/AC:M/Au:N/C:C/I:C/A:C)}\\
            \multicolumn{1}{p{.85\textwidth}}{{\bf Generated Question}: What is the CVSS score for the vulnerability in WebSphere Application Server?}  \\
            \multicolumn{1}{p{.85\textwidth}}{{\bf Generated Answer}: CVSS Base Score 9.3.} 
        } \\
        \midrule
        \multicolumn{1}{p{.15\textwidth}}{\textbf{RADA (Ours)}} & 
        \makecell{
            \multicolumn{1}{p{.85\textwidth}}{{\bf Context}: The Wii Homebrew Applications (such as The Homebrew Channel, Emulators, Games, Media players, etc.), USB Loader GX's NAND emulation feature and DIOS MIOS (Lite) require a FAT32 partition format. You can use a FAT32 SD/SDHC card for that purpose and set your USB hard drive as NTFS or Ext partition format.}\\
            \multicolumn{1}{p{.85\textwidth}}{{\bf Generated Question}: What partition format is required for certain Wii Homebrew applications?}  \\
            \multicolumn{1}{p{.85\textwidth}}{{\bf Generated Answer}: FAT32} 
        } \\
        \bottomrule
        \end{tabular}
    }
\end{table*}

\end{document}